\documentclass[runningheads]{llncs}

\usepackage{eccv}

\usepackage{eccvabbrv}

\usepackage{graphicx}
\usepackage{booktabs}
\usepackage{multirow}
\usepackage{makecell}

\usepackage[width=122mm,left=12mm,paperwidth=146mm,height=193mm,top=12mm,paperheight=217mm]{geometry}

\usepackage[accsupp]{axessibility}  

\usepackage{hyperref}

\usepackage{orcidlink}

\begin{document}
\title{Make Your Actor Talk: Generalizable and High-Fidelity Lip Sync with Motion and Appearance Disentanglement}

\titlerunning{MyTalk}

\author{Runyi Yu$^{1}$ \and
Tianyu He \and
Ailing Zhang$^{1}$ \and
Yuchi Wang$^{1}$ \and
Junliang Guo \and
Xu Tan \and \\
Chang Liu$^{3}$ \and
Jie Chen$^{1,2}$ \and
Jiang Bian}
\authorrunning{Runyi Yu et al.}

\institute{
$^{1}$Peking University \quad $^{2}$Peng Cheng Laboratory \quad $^{3}$Tsinghua University \\
\vspace{1mm}
\email{ingrid\_yu@stu.pku.edu.cn}\\
\vspace{1mm}
\url{https://Ingrid789.github.io/MyTalk/}}

\maketitle

\vspace{-10mm}
\begin{figure}
  \centering
  \includegraphics[height=11.8cm]{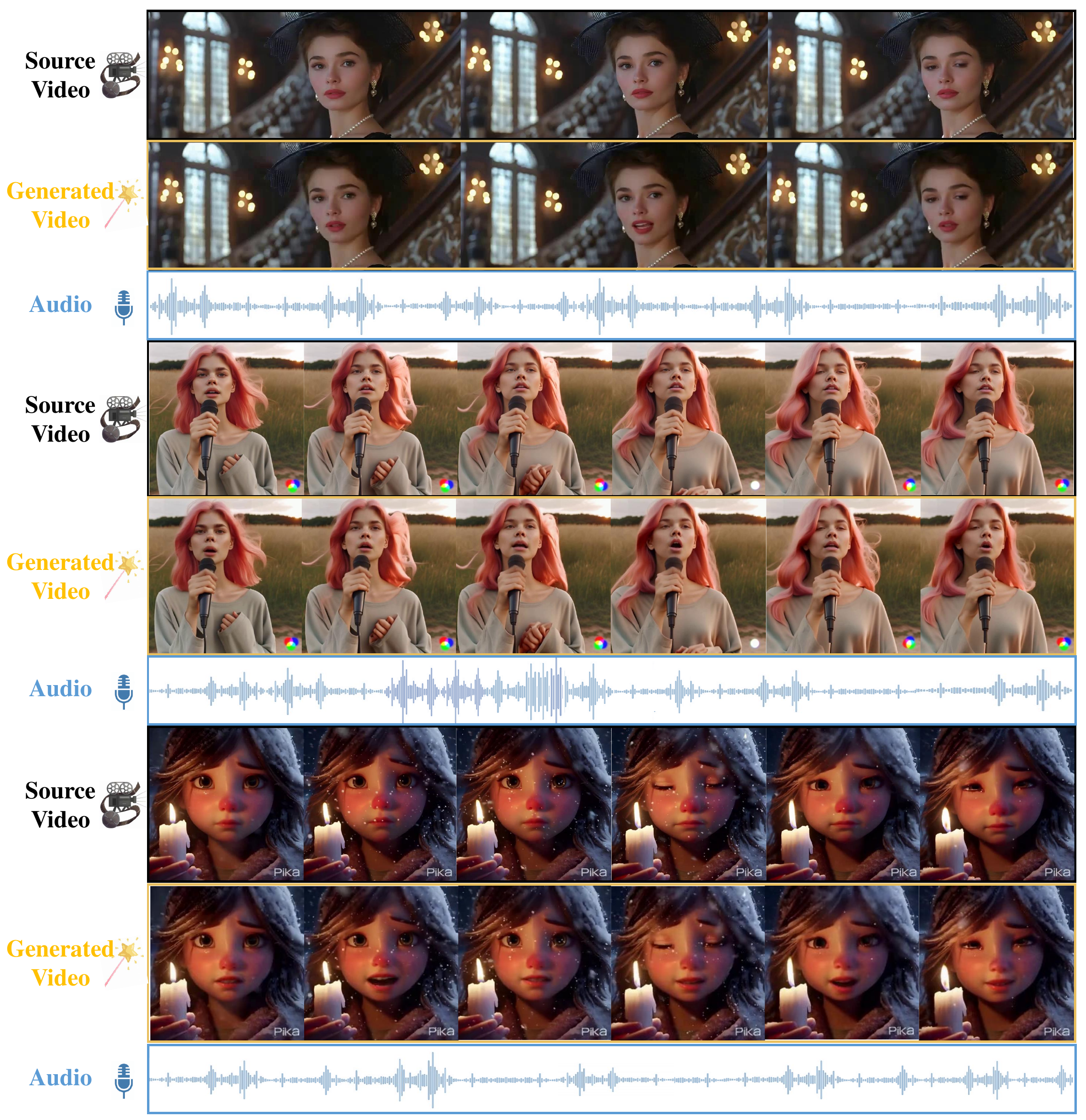}
  \caption{Samples of the original talking videos and the ones generated by our method according to the given speech input, showcasing that our method performs well in both lip sync and visual detail preservation. It also generalizes well to the unknown and out-of-domain characters (e.g., the bottom case), enabling seamless lip sync for AI-generated videos (e.g., the middle and bottom cases).}
  \label{fig1:introduction}
\end{figure}


\begin{abstract}
  We aim to edit the lip movements in talking video according to the given speech while preserving the personal identity and visual details. The task can be decomposed into two sub-problems: (1) speech-driven lip motion generation and (2) visual appearance synthesis. Current solutions handle the two sub-problems within a single generative model, resulting in a challenging trade-off between lip-sync quality and visual details preservation. Instead, we propose to disentangle the motion and appearance, and then generate them one by one with a speech-to-motion diffusion model and a motion-conditioned appearance generation model. However, there still remain challenges in each stage, such as motion-aware identity preservation in (1) and visual details preservation in (2). Therefore, to preserve personal identity, we adopt landmarks to represent the motion, and further employ a landmark-based identity loss. To capture motion-agnostic visual details, we use separate encoders to encode the lip, non-lip appearance and motion, and then integrate them with a learned fusion module. We train MyTalk on a large-scale and diverse dataset. Experiments show that our method generalizes well to the unknown, even out-of-domain person, in terms of both lip sync and visual detail preservation. We encourage the readers to watch the videos on our \hyperlink{https://Ingrid789.github.io/MyTalk/}{project page}.
\end{abstract}

\label{sec:intro}
\section{Introduction}
\begin{figure}[t]
  \centering
  \includegraphics[height=8cm]{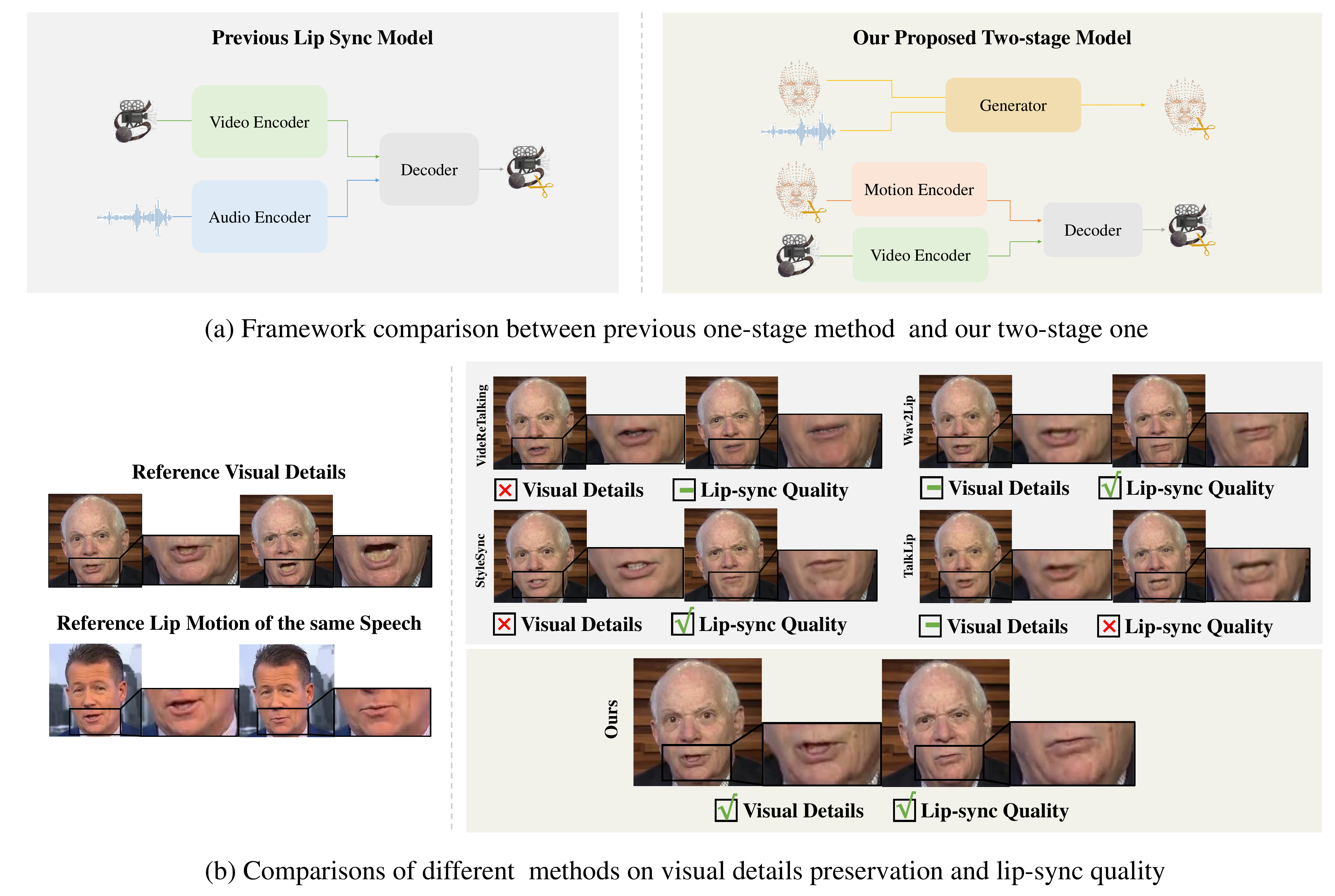}
  \caption{A brief illustration of previous methods and ours. We compare the edited talking videos generated by different models to evaluate their visual details preservation and lip-sync quality. (a) gives a straightforward framework comparison. (b) shows the evaluation of different methods of visual detail preservation and lip-sync quality. Previous one-stage methods all struggle to simultaneously preserve visual details and ensure lip-sync quality, while our proposed motion-appearance disentangled two-stage method achieves excellent results in both aspects.}
  \label{fig1.5:limitations}
\end{figure}

Given a video that includes talking individuals, speech-driven talking video lip sync re-generates the lip motion in line with the speech content while preserving the personal identity and appearance visual details~\cite{thies2020neural, prajwal2020lip}. It has huge application prospects in industries like education, live streaming, etc. Recently, with the significant advancements made in AI-generated videos~\cite{guo2023animatediff, blattmann2023stable, chen2023videocrafter1, kondratyuk2023videopoet}, it also shows great potential in editing the lip motion of the generated actors as shown in Fig.~\ref{fig1:introduction}.

Due to the significant domain gap between visual and speech content, generating visual content that adheres to the visual identity and speech content at the same time is difficult. Early attempts train or adapt a specific model for each person~\cite{thies2020neural, tang2022memories, du2023dae, guo2021ad}, where the identity information is implicitly memorized in the model weights. Although high-quality synthesis is achieved, the trained model can not be employed for unknown persons, thus limiting its broader application. Several studies tackle this challenge by learning the identity and facial appearance in an in-content manner~\cite{prajwal2020lip, cheng2022videoretalking, guan2023stylesync, wang2023seeing, shen2023difftalk}. For example, Wav2Lip~\cite{prajwal2020lip} masks the lower-half face of the input video and generates the masked region conditioned on the input speech and the reference frames. By training on abundant identities, the model learns to infer the masked facial appearance from the given reference frames. Along this line, the following works further improve the diversity, lip movements and visual quality~\cite{guan2023stylesync,wang2023seeing,shen2023difftalk,cheng2022videoretalking}, etc. However, the methods above all follow a one-stage framework: the speech-driven lip motion generation and identity-aware appearance generation are handled within a single model, and the vision-speech alignment difficulty makes these methods a trade-off between lip-sync quality and visual details. Fig~\ref{fig1.5:limitations} gives the visual illustration. 

In this work, we observe that the motion depicted by landmarks~\cite{wood2021fake} can serve as a mediator reducing the disparity between the visual and speech domains.
Accordingly, we propose to disentangle the motion and appearance in talking video, and generate them one by one in a two-stage framework. This insight is also inspired by previous work on talking video generation with a single portrait image~\cite{he2023gaia}, however, there still remain challenges for each stage: First, it is crucial to preserve the personal identity during the motion generation, such as ensuring consistency of face contours between the original and the generated one. 
Second, it is non-trivial to generate person-specific details in the lip region while preserving the dynamic non-lip region in the original video. To realize appearance and motion disentanglement and overcome the challenges, we present \textbf{M}ake \textbf{Y}our Actor \textbf{Talk}, abbreviated as \textbf{MyTalk}. It consists of two models: a speech-to-motion diffusion model that generates facial landmarks~\cite{wood2021fake} according to the input speech, and a motion-conditioned appearance generation model that synthesizes the output video conditioned on the generated landmarks and reference frames. To tackle the above two challenges, for the speech-to-motion diffusion model, we introduce the extracted identity feature as an extra condition and employ a landmark-based identity loss to preserve personal identity. For the motion-conditioned appearance generation model, since the reference frames are with different motions, to capture motion-agnostic visual details, we use three encoders: a lip appearance encoder, a non-lip appearance encoder, and a landmark encoder to encode the appearance and motion representations. The learned representations are fused and then decoded to output videos using the decoder.

To enable better generalization, we train MyTalk on a large-scale and diverse dataset that consists of 15K unique identities. Experimental results demonstrate that MyTalk achieves excellent results in visual detail preservation and lip-sync quality. Owing to the superiority of the disentangled modeling, it also generalizes well to the unknown characters in unseen domains (see Fig.~\ref{fig1:introduction}), and enables interesting appearance editing and emotion control on the original video as shown in Fig.~\ref{fig3:appearance_edit}. To conclude, our contributions can be summarized as:

\begin{itemize}
    \item We leverage motion (i.e., landmarks) to reduce the disparity between the video and speech content domains, and further model the talking video lip sync into two sub-problems: speech-driven motion generation and motion-conditioned visual appearance synthesis.
    \item We propose MyTalk, a new two-stage lip sync method, which disentangles the motion and appearance and generates them step-by-step with separate models. 
    \item To preserve personal identity, we introduce the extracted identity feature as an extra condition and employ a landmark-based identity loss. To capture motion-agnostic visual details, we use separate encoders to encode the motion, lip and non-lip appearance, and combine them with a learnable fusion module.
    \item With the scaled training data, MyTalk generalizes well to the unknown and out-of-domain characters, it also shows great controllability in terms of lip appearance and talking emotions.
\end{itemize}

\section{Related Work}
Talking video generation aims to generate talking videos with lip movements synchronized with the given speech~\cite{prajwal2020lip, cheng2022videoretalking, guan2023stylesync, he2023gaia, zhang2023sadtalker, guo2021ad, tang2022memories}. There are two main subdivisions in this field, including talking video generation and lip sync. 

\subsection{Speech-driven Talking Video Generation}
Talking video generation aims to generate lifelike talking videos according to a speech and single portrait reference image. This requires the model to generate consistent identity, natural head pose and high-quality lip motion. 
Various works leverage structural information as the intermediate representation such as landmarks~\cite{he2023gaia, chen2019hierarchical, zhou2020makelttalk, wang2024instructavatar}, 3D Morphable Models~\cite{blanz2023morphable, zhang2021flow, zhang2023sadtalker, ren2021pirenderer} and 3D meshes~\cite{chen2020comprises}. Since structural information is still coupled with complex information, these methods typically require additional processing to ensure effective generation, such as MakeItTalk~\cite{zhou2020makelttalk} decompose the content and speaker information from the speech; PC-AVS~\cite{zhou2021pose} and GAIA~\cite{he2023gaia} disentangles the head pose and facial expression. Recently, NeRF~\cite{guo2021ad, shen2022learning} is also introduced in talking video generation, which requires person-specific training and performs poorly across identity.

In this work, we take inspiration from GAIA~\cite{he2023gaia} to leverage structural information and introduce it into the talking video lip sync. We also propose novel techniques to handle personal identity and visual detail preservation.

\subsection{Speech-driven Talking Video Lip Sync}
Speech-driven talking video lip sync aims to generate the edited talking video with lip movements re-synchronized to the speech and others, e.g. head motion, identity appearance and background dynamics, remaining identical to the given video. The challenges mainly focus on visual detail preservation and lip sync. Some works~\cite{guo2021ad, wen2020photorealistic, thies2020neural} train or fine-tune a specific model for each identity, which achieves impressive results in identity visual details and lip-sync quality. However, the specific trained models cannot generalize across different identities. Other methods adopt an end-to-end framework and show great robustness in personal identity. Before the diffusion emerges, GANs~\cite{das2020speech, gu2020flnet, prajwal2020lip} are usually used as the primary techniques. LipGAN~\cite{kr2019towards} is the first commonly known GAN-based method, and Wav2Lip~\cite{prajwal2020lip} boosts the lip-sync quality by introducing a pre-trained discriminator. Similarly, the latter work ~\cite{wang2023seeing} adopts a lip-reading expert to further elevate lip sync. VideoReTalking~\cite{cheng2022videoretalking} incorporates several pre-processing and post-processing that significantly enhance the visual quality, and StyleSync~\cite{guan2023stylesync} strengthens the personalized generation. A recent work DiffTalk~\cite{shen2023difftalk} tries to edit the talking video in latent space, and models the generation in a diffusion way.

We note that all these generalized methods handle lip motion and appearance generation within a single model. However, as visualized in Fig.~\ref{fig1.5:limitations}, these one-stage models struggle to simultaneously process the visual and speech content, resulting in a trade-off between visual details preservation and lip-sync quality.

\section{Method}
\subsection{Preliminary}
\subsubsection{Problem Definition.} Given a talking video $x$ and a speech sequence $s$, talking video lip sync aims to generate video $\hat{x}$ that lip synchronized with speech $s$ and the dynamic non-lip regions adhere to the original video $x$. 

\begin{figure}[t]
  \centering
  \includegraphics[height=7.8cm]{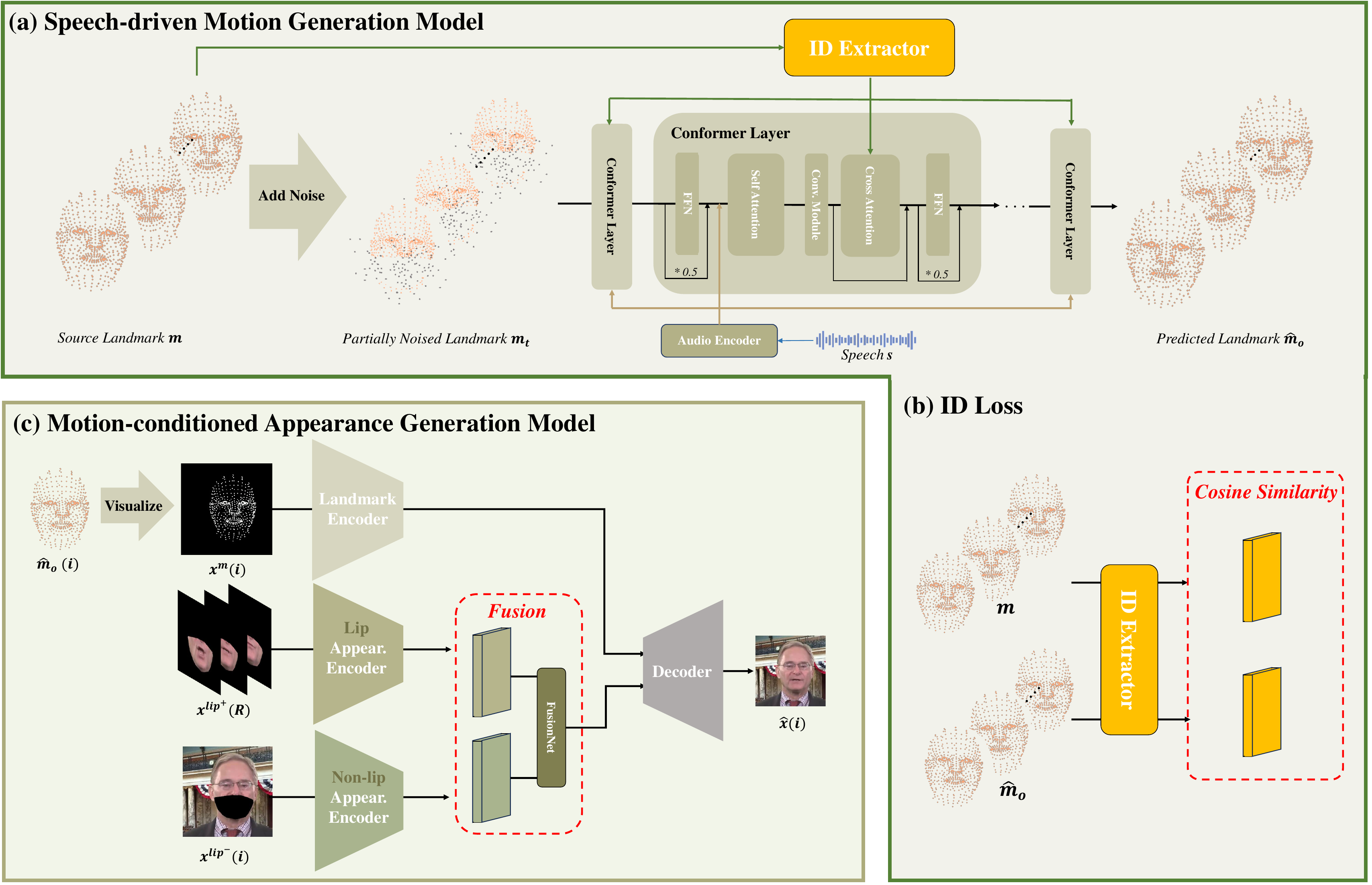}
  \caption{Our proposed MyTalk adopts a motion-appearance disentangled two-stage framework to realize talking video lip sync. (a) In the first stage, we adopt a speech-driven motion generation model to generate motion (i.e., landmark) sequences from the input speech with the diffusion model. (b) To better preserve the motion identity, we design an identity extractor and the corresponding identity loss in the motion generation model. (c) In the second stage, we use separate encoders to encode the motion-agnostic lip, non-lip appearance, and the generated motion. The encoded representations are fused with a FusionNet and decoded to the output video.
  }
  \label{fig2:framework}
\end{figure}

\subsubsection{Previous Solutions.}
\label{method:previous_work}
Previous works typically use a one-stage model to edit the lip movements in talking videos, as shown in the right part in Fig.~\ref{fig1.5:limitations} (a). These methods can be formulated as:
\begin{equation}
\label{eq:one-stage}
       \hat{x}=\mathcal{D}(\mathcal{E}_{X}(x), \mathcal{E}_{S}(s)),
\end{equation}
where $\mathcal{E}_{X}$ and $\mathcal{E}_{S}$ represent the video and speech encoder, and $\mathcal{D}$ represents the decoder. Owing to the significant domain gap between visual $x$ and speech $s$ signals, these one-stage models struggle to align the vision and speech features, resulting in the loss of visual details or inferior lip-sync quality.

\subsection{Overview}
\label{method_overview}

To reduce the disparity between the visual and speech content, we propose MyTalk which decomposes the talking video lip sync into two stages as shown in Fig~\ref{fig2:framework}. In the first stage, we adopt dense landmarks~\cite{wood2021fake} as the motion representation and present a speech-driven motion generation model that generates landmarks according to the input speech. The generated landmarks are then represented as 2D images and leveraged in the second stage. To preserve personal identity, we introduce the extracted identity features as an extra condition and employ a landmark-based identity loss. In the second stage, we propose a motion-conditioned appearance generation model that synthesizes the video according to the landmark images and reference frames. Since our goal is to generate the lip region while preserving the dynamic non-lip region, we encode lip, non-lip appearance and motion with three separate encoders. The encoded features are then fused with a FusionNet followed by a decoder to produce output video.



\subsection{Speech-driven Motion Generation}
\label{method_motion_model}

Given the input speech, the speech-driven motion generation model is expected to predict the aligned motion corresponding to the speech content. To achieve this, we use the dense landmark~\cite{wood2021fake} as the motion representation and extract the landmarks $m$ from the given talking video as the input information. For the input speech, we extract its feature using a pre-trained wav2vec model~\cite{schneider2019wav2vec}. After obtaining the original landmarks and speech feature, we employ a diffusion model~\cite{ho2020denoising, wang2023unlimited, wang2022zero} with Conformer backbone~\cite{gulati2020conformer} to generate the motion sequence conditioned on the speech feature and identity information extracted from reference landmarks. It should be noted that, since our goal is to re-generate the lip motion while preserving the dynamics of the non-lip regions, we also make the generation conditioned on the landmarks of the upper face $m^{nl}$ and only yield the landmarks of the lower face $m^{l}$.

\subsubsection{Diffusion Model.}
In the forward diffusion process, the landmarks of the lower face $m^{l}$ undergo a progressive noising process, making it into Gaussian noise over an infinite number of time steps. The noising process can be formulated as:
\begin{equation}
\label{eq:diffusion}
\begin{split}
    q(m_{t}^{l} | m_{t-1}^{l})=\mathcal{N}(\sqrt{\alpha_t}m_{t-1}^{l}, (1-\alpha_t)I),
\end{split}
\end{equation}
where $\alpha_t\in(0,1)$ is a constant hyper-parameter, and $m_{0}^{l}$ is drawn from the data distribution. When $\alpha_t$ is small enough, we can approximate $m_{t}^{l}\sim\mathcal{N}(0, I)$.

Subsequently, in the reverse diffusion phase, the model learns the distribution $p(m_{0}^{l}|C)$ and reconstructs the original landmark $\hat{m}_{0}^{l}$ by denoising the $m_{t}^{l}$. In practice, to ensure the smoothness between the upper face and the generated lower face, we add the landmarks of the upper face $m^{nl}$ of the current frame as the prefix of Conformer input. Therefore, each reverse diffusion step can be formulated as:
\begin{equation}
\label{eq:diffusion1}
\begin{split}
    \hat{m}_{0}^{l}=\mathcal{G}(m_{t}^{l}|C, m^{nl}),
\end{split}
\end{equation}
where $\mathcal{G}$ represents the Conformer backbone, and $C$ represents the conditions (i.e., the speech feature, the diffusion time step and identity embedding extracted from reference landmarks). We concatenate the generated landmarks of the lower face $\hat{m}_{0}^{l}$ with the given landmarks of the upper face $m^{nl}$ to formulate the motion output $\hat{m}_{0}$.


\subsubsection{Identity Preservation.}
\label{method_gen_identity}

Since the landmarks express facial contours that convey identity information in the form of facial shapes. Owing to the lack of identity constraint, the motion generation model may struggle to produce identity-aware landmark sequences solely with the reconstruction loss. The insight here is that the reconstruction objective is a low-level constraint for landmarks, while identity preservation is a high-level target, which calls for a corresponding high-level constraint. To this end, we first train a facial recognition model (i.e., the identity extractor $\mathcal{M}$) that takes the facial landmarks as input and outputs the facial embeddings. This model is composed of several MLP layers and trained with the ArcFace~\cite{deng2019arcface} loss, which facilitates the embeddings that belong to the same person to be close. Then, we integrate the pre-trained identity extractor into the motion generation model in two ways: (1) We regard the embeddings extracted from the reference landmarks as the condition during the generation. (2) We employ an identity loss that minimizes the embeddings extracted from the reference landmarks and the generation landmarks.



\subsubsection{Conditioning.} Our conditions $C$ consist of the speech feature $z^s$, the noise time step $t$, and the identity embedding $z^{id}$.
To give the conditions to the diffusion model, for noise time step $t$, we project it into a sinusoidal encoding and further add it to the input of each Conformer layer. For speech $z^s$ and identity $z^{id}$ feature, we first project them into the Conformer dimensions with MLP. Subsequently, we add the projected $z^s$ to the hidden feature of each Conformer block in an element-wise manner and integrate the projected $z^{id}$ into the Conformer block via cross-attention.


\subsubsection{Training.} We expect our diffusion model to reconstruct the motion sequence under the speech condition and simultaneously ensure consistency of personal identity. Hence, the denoising process can be optimized with the following objective:
\begin{equation}
\label{eq:diffusion_1}
\begin{split}
    L_{DM}=\mathbb{E}_{m,C}\bigg[ \|\hat{m}_0 - m\|_{2}^{2}\bigg] + \mathbb{E}\bigg[ \|1-\hat{z}^{id} \cdot {z^{id}}^T\| \bigg],
\end{split}
\end{equation}
where $C=\{t,z^s,z^{id}\}$ represents the conditions. $\hat{z}^{id}$ and ${z^{id}}$ are the identity embedding obtained with identity extractor $\mathcal{M}$.

\subsection{Motion-conditioned Appearance Generation}
\label{method_VAE}

Given the generated landmarks $\hat{m}$ from the motion generation model, the appearance generation model is responsible for producing the output video conditioned on the generated landmarks and reference frames. Ideally, it should preserve the visual details of the reference frames and generate talking videos that are synchronized with the landmarks $\hat{m}$. To this end, we first represent the landmarks in an image way by projecting the coordinates of landmarks to the image coordinate and obtain $x^m$. After that, inspired by GAIA~\cite{he2023gaia}, we leverage a Variational AutoEncoder (VAE)~\cite{kingma2013auto} to synthesize the appearance. However, different from GAIA which generates talking video with a single frame, our goal is to generate the lip region while preserving the dynamic non-lip region. Therefore, we encode the lip region and non-lip region separately, where the non-lip region comes from the given video and the lip region should be generated according to the generated landmarks. Specifically, we employ three encoders: a lip appearance encoder $\mathcal{E}_{R}$, a non-lip appearance encoder $\mathcal{E}_{A}$ and landmark encoder $\mathcal{E}_{M}$. The encoded lip and non-lip appearance features are then fused with a FusionNet $\mathcal{F}$. Finally, the decoder $\mathcal{D}$ takes all features and produces the output video.

\subsubsection{Appearance and Motion Encoding.} 
We denote a single video frame as $x(i)$, and its corresponding projected landmark image as $x^{m}(i)$. With $m(i)$, we can identify the lip and non-lip regions, which enables us to produce lip and non-lip region masked video frame, denoted as $x^{l}(i)$ and $x^{nl}(i)$, respectively. 

To mitigate the risk of missing lip-related details in a single-frame reference, such as occluded teeth information in lip closed frame, we propose a multi-frame reference strategy. Specifically, we randomly select three frames from the same video clip as reference images and get the non-lip region masked out, denoted as $x^{l}(R)$. For the current frame $x(i)$, we use three encoders to obtain the corresponding features:
\begin{equation}
    \begin{split}
       z^{l}(R)&=\mathcal{E}_{R}(x^{l}(R)), \\
       z^{nl}(i)&=\mathcal{E}_{A}(x^{nl}(i)), \\
       z^{m}(i)&=\mathcal{E}_{M}(x^m(i)),
    \end{split}
\end{equation}
where $z^{l}(R)$, $z^{nl}(i)$ and $z^{m}(i)$ are the encoded features. We use similar model architecture for $\mathcal{E}_{R}$, $\mathcal{E}_{A}$ and $\mathcal{E}_{M}$, which is demonstrated in detail in the Appendix.

\subsubsection{FusionNet.} Since $z^{l}(R)$ and $z^{nl}(i)$ both represent appearance information, we propose to use FusionNet to integrate them into one feature that contains both lip and non-lip appearance:
\begin{equation}
\label{eq:FusionNet}
       z^{a}(i)=\mathcal{F}(z^{l}(R), z^{nl}(i)),
\end{equation}
where $z^{a}(i)$ represents the fused appearance feature. FusionNet $\mathcal{F}$ adopts convolutions to fuse the appearance latent, and its detailed architecture is depicted in the Appendix. 

\subsubsection{Decoding.} We utilize a decoder $\mathcal{D}$ to recover the visual details from appearance latent $z^{a}(i)$ and lip movements from motion latent $z^{m}(i)$. These latent features are concatenated together in channel dimension  and used for decoding:
\begin{equation}
\label{eq:decoder0}
       \hat{x}(i)=\mathcal{D}(z^{a}(i), z^{m}(i)).
\end{equation}

\subsubsection{Training.} Our appearance generation model is trained with the L1 reconstruction loss. Since the model handles a generative task, we incorporate a VAE training objective with a KL-penality on the latent. 
Building on prior studies~\cite{esser2021taming,rombach2022high,he2023gaia}, we incorporate a discriminator loss to further improve the generation visual quality. Specifically, the discriminator $\mathcal{D}_{disc}$ takes the real frame $x(i)$ and reconstructed frame $\hat{x}(i)$ as the input and computes the discriminator loss $L_D$:
\begin{equation}
\label{eq:decoder1}
       L_{disc}(x(i), \hat{x}(i))=log\mathcal{D}_{disc}(x(i))+log(1-\mathcal{D}_{disc}(\hat{x}(i))).
\end{equation}
Adding the reconstruction loss $L_{rec}$ and KL-penality $L_{KL}$, we can get our final loss function as:
\begin{equation}
\label{eq:decoder2}
    L_{VAE}=\min_{P} \max_{\mathcal{D}_{disc}} (L_{rec}(x; P) + L_{KL}(x; P) + L_{disc}(x; \mathcal{D}_{disc})),
\end{equation}
where $P$ denotes the set of parameters within $\mathcal{E}_{R}, \mathcal{E}_{A}, \mathcal{E}_{M}, \mathcal{F},$ and $ \mathcal{D}$. 

\subsection{Inference}
Our method incorporates two models: a speech-driven diffusion model to generate motion and a motion-conditioned appearance generation model to generate the edited talking video. 
During inference, we obtain the reference motion landmark $m$ from the source video $x$. Then for the lower face, we start from Gaussian noise and infer the motion sequence $\hat{m}$ conditioned on the input speech, the identity embedding and the obtained landmarks of the upper face with the denosing process~\cite{ho2020denoising}. Subsequently, the generated motion sequence $\hat{m}$, the lip-masked source video $x^{nl}$ and the reference frames $x^{l}(R)$ are fed together into the appearance generation model to generate the final talking video $\hat{x}$.

\section{Experiments}
Comprehensive experiments are conducted to evaluate our proposed MyTalk. We organize the experiments into three parts. Firstly, in Sec.~\ref{main_exp}, we conduct primary experiments to evaluate the generation quality of different models with both objective and subjective assessments. To better measure the lip-sync quality, we conduct additional experiments on generation with random speech input, which evaluates the generated lip-sync quality conditioned on unaligned video-speech pairs. Subsequently, we compare our appearance generation model with the state-of-the-art one. Finally, we conduct elaborate ablation studies on both our motion and appearance generation model, which is demonstrated in Sec.~\ref{ablation_study}.

\vspace{-5mm}
\begin{table}[t]
  \caption{\textbf{Objective comparisons of our model with previous lip sync baselines.} MyTalk achieves superior results in visual and lip-sync quality. The state-of-the-art results on visual quality and ID similarity metrics showcase our superiority in preserving visual details.}
  \label{tab1:obj_align}
  \centering
  \begin{tabular}{@{}lcccccccc@{}}
    \toprule
    \multirow{2}{*}{Method} & \multicolumn{7}{c}{Objective Evaluation} \\
    \cmidrule(l){2-8}
    & PSNR$\uparrow$ & SSIM$\uparrow$ & LPIPS$\downarrow$ & FID$\downarrow$ & Sync$_{dist}\downarrow$ & Sync$_{conf}\uparrow$ & ID$_{sim}\uparrow$ \\
    \midrule
    Wav2Lip~\cite{prajwal2020lip} & 30.967 & 0.894 & \underline{0.095} & 7.703 & \textbf{6.249} & \textbf{7.829} & 0.921 \\
    VideoReTalking~\cite{cheng2022videoretalking} & 29.628 & 0.885 & 0.101 & 9.628 & 6.978 & 6.986 & 0.895 \\
    TalkLip~\cite{wang2023seeing} & \underline{31.140} & \underline{0.896} & \underline{0.095} & \underline{7.401} & 8.107 & 5.300 & \underline{0.926} \\
    \midrule
    MyTalk (ours) & \textbf{33.547} & \textbf{0.906} & \textbf{0.091} & \textbf{5.142} & \underline{6.544} & \underline{7.461} & \textbf{0.947} \\
    \bottomrule
  \end{tabular}
\end{table}

\begin{table}[t]
    \centering
        \begin{tabular}[t]{ccc}
        \vspace{-0mm}
        \begin{minipage}[t]{.45\linewidth}
            \centering
            \scriptsize
            \captionof{table}{\textbf{Objective comparisons on unpaired test set.} 
            As shown here, our method has superior and consistent performance.}
            \label{tab:obj_unalign}
            \begin{tabular}{@{}l|cc}
    		\toprule[1.0pt]
    		\multirow{2}{*}{Method} & \multicolumn{2}{c}{Objective Evaluation} \\
                \cmidrule(l){2-3}
                 &  Sync$_{dist}\downarrow$ &  ID$_{sim}\uparrow$ \\
    		\midrule
                Wav2Lip~\cite{prajwal2020lip} & \underline{6.711} & 0.916 \\
                VideoReTalking~\cite{cheng2022videoretalking} & 7.580 & 0.694 \\
    		  TalkLip~\cite{wang2023seeing} & 9.321 & \underline{0.922} \\
                \midrule
    		  MyTalk (ours) & \textbf{6.703} & \textbf{0.943} \\
    		\bottomrule[1.0pt]
    	\end{tabular}
        \end{minipage} &
        \begin{minipage}[t]{.50\linewidth}
            \centering
            \scriptsize
            \captionof{table}{\textbf{Subjective comparisons.} The metrics reported here specifically refer to: visual quality, lip-sync quality, visual details preservation and identity consistency.}
            \label{tab:subj}
            \begin{tabular}{@{}l|@{}cccc@{}}
    		\toprule[1.0pt]
    		\multirow{2}{*}{Method} & \multicolumn{4}{c}{Subjective Evaluation} \\
                \cmidrule(l){2-5}
                & Vis.$\uparrow$ & Sync$\uparrow$ & Detail$\uparrow$ & Consis.$\uparrow$ \\
    		\midrule
                Wav2Lip~\cite{prajwal2020lip} & 3.04 & 3.32 & 2.98 & 3.74 \\
                VideoReTalking~\cite{cheng2022videoretalking} &  3.83 & 3.42 & 3.55 & 3.78 \\
    		  TalkLip~\cite{wang2023seeing} & 3.35 & 3.46 & 3.40 & 3.90 \\
                \midrule
    		  MyTalk (ours) & \textbf{4.22} & \textbf{3.94} & \textbf{4.30} & \textbf{4.45} \\
    		\bottomrule[1.0pt]
    	\end{tabular}
        \end{minipage} &
        \vspace{0mm}
        \end{tabular}
\end{table}

\subsection{Experimental Settings}
\label{main_exp}
\subsubsection{Dataset.} 
Our training data comes from public datasets, including HDTF~\cite{zhang2021flow} and CC v1 $\&$ v2~\cite{hazirbas2021towards, porgali2023casual}, and internal collected datasets. Our dataset varies from gender, race, age, and language. The whole dataset includes $16$K unique identities and has a cumulative duration of $1.1$K hours. All the videos in the dataset are cropped and processed by following previous work~\cite{he2023gaia} with a $256\times256$ resolution. We randomly sample $30$ identities covering 2 hours as the validation set and leaving the remains as the training set.

To eliminate the potential training overlap and fairly evaluate the generative ability of each method, we sample $507$ videos (around 74K frames in total) from TalkingHead-1KH~\cite{wang2021one} dataset, which comprises $170$ identities and spans a wide range of languages. To conduct diverse tests, we formulate two test sets from the sampled $507$ videos: a paired test set with the original video and speech, and an unpaired test with randomly sampled speech. By default, we conduct all the comparison experiments on the paired test set. We also provide comparisons with baseline methods in the unpaired test set.

\begin{table}[t]
    \centering
    \begin{tabular}[t]{ccc}
    \vspace{-5mm}
    \begin{minipage}[t]{.49\linewidth}
        \centering
        \captionof{table}{\textbf{Comparisons of our appearance generation model with baseline.} See Sec.~\ref{main_exp} for details.}
        \label{tab:gaia_compare}
        \resizebox{\columnwidth}{!}{
         \begin{tabular}{@{}l|cccc@{}}
         \toprule[1.0pt]
             Method & PSNR$\uparrow$ & SSIM$\uparrow$ & LPIPS$\downarrow$ & FID$\downarrow$ \\
             \midrule
             GAIA~\cite{he2023gaia} & 24.205 & 0.757 & 0.175 & 14.568 \\
             MyTalk (ours) & \textbf{34.153} & \textbf{0.926} & \textbf{0.092} & \textbf{4.224} \\
         \bottomrule[1.0pt]
         \end{tabular}
          }
    \end{minipage} &
    \begin{minipage}[t]{.50\linewidth}
            \centering
            \scriptsize
            \captionof{table}{\textbf{Ablation study on condition types in motion generation model.}  Sec.~\ref{ablation_study} gives detailed explanation.}
            \label{tab:diff_ablation}
            \begin{tabular}{l|cc}
    		\toprule[1.0pt]
    		Method & Sync$_{dist}\downarrow$ & ID$_{sim}\uparrow$ \\
                \midrule		
                Ours (ID embed. cross-atten.) & 6.544 & 0.947 \\
                \midrule
    		concat. key landmark &  6.718 & 0.923 \\
    		add key landmark & 7.509 & 0.932 \\
                key landmark cross-attn. & 6.814 & 0.914 \\
    		\bottomrule[1.0pt]
    	\end{tabular}
        \end{minipage}
    \vspace{0mm}
    \end{tabular}
\end{table}

\subsubsection{Metric.} 
We utilize various metrics including objective and subjective ones to provide an overall evaluation of our method. 
\begin{itemize}
    \item \textbf{Objective Metrics.} We adopt the objective metrics used in previous studies~\cite{prajwal2020lip, guan2023stylesync, he2023gaia, shen2023difftalk, cheng2022videoretalking}, including metrics to evaluate visual quality, lip sync accuracy and identity consistency. Specifically, we report commonly used PSNR, SSIM~\cite{wang2004image}, LPIPS~\cite{zhang2018unreasonable} and FID~\cite{heusel2017gans} to evaluate the generated visual quality. We adopt SyncNet~\cite{chung2017out} distance and conference scores, denoted as Sync$_{dist}$ and Sync$_{conf}$, to measure the lip sync quality. Inspired by ~\cite{guan2023stylesync}, we leverage the ArcFace~\cite{deng2019arcface} network to compute the feature cosine similarity, denoted as ID$_{sim}$, between the ground truth and generated video frames, and use it to check the identity preservation ability.
    \item \textbf{Subjective Metrics.} We adopt four subjective metrics including visual quality, lip-sync quality, visual details preservation and identity consistency. In our user studies, we invite 20 participants to rate the talking videos generated by different methods. Participants are asked to rate a total of 44 videos on a 1-5 scale, with the average scores being reported.
\end{itemize}

\subsubsection{Implementation Details.} The identity extractor, motion generation model, and appearance generation model are trained separately. We adopt the Adam~\cite{kingma2014adam} optimizer for each model training. We perform all the experiments on 8 V100 GPUS. The identity extractor is trained for 100 epochs with a basic learning rate of $1e^{-6}$. The speech-driven motion generation model is trained for 3.5K epochs with an inverse square root learning rate schedule, and its basic learning rate is $1e^{-4}$. The appearance generation model is trained for 10 epochs on 256$\times$256 resolution with a stable learning rate of $1e^{-6}$. The landmark sequence is padded to have a uniform length of 250 frames.

\subsubsection{Results.}
We compare our proposed MyTalk with previous methods, including Wav2Lip~\cite{prajwal2020lip}, VideoReTalking~\cite{cheng2022videoretalking} and TalkLip~\cite{wang2023seeing}, with both objective (in Tab.~\ref{tab1:obj_align} and Tab.~\ref{tab:obj_unalign}) and subjective (in Tab.~\ref{tab:subj}) evaluations. It can be observed that MyTalk surpasses all the baselines by a large margin in terms of visual quality and identity similarity, showcasing our great success in visual details preservation. Specifically, as shown in Fig.~\ref{fig1.5:limitations} (b), other methods tend to lose the original lip appearance or produce a blurred one, whereas our approach effectively preserves the lip details. Moreover, MyTalk consistently delivers superior lip-sync quality with both paired and unpaired test sets. However, comparing the lip-sync quality in Tab.~\ref{tab:obj_unalign}, the performance of other methods all shows an obvious decline when handling the unpaired test set. Our leading results in Tab.~\ref{tab:subj} further demonstrate the effectiveness of MyTalk in visual details preservation and lip-sync quality. In addition, we compare our appearance generation model with the GAIA~\cite{he2023gaia}, and the results in Tab.~\ref{tab:gaia_compare} affirm our effectiveness in terms of visual quality.

\subsection{Ablation Study}
\label{ablation_study}
\begin{table}[t]
    \centering
        \begin{tabular}[t]{ccc}
        \vspace{-0mm}
        \begin{minipage}[t]{.48\linewidth}
            \centering
            \captionof{table}{\textbf{Ablation study of loss design in motion generation model.} See Sec.~\ref{ablation_study} for details.}
            \label{tab:loss_ablation}
            \begin{tabular}{l|cc}
    		\toprule[1.0pt]
    		Method & Sync$_{dist}\downarrow$ & ID$_{sim}\uparrow$ \\
                \midrule
                Ours (ID loss $\times$ 1.0) & 6.544 & 0.947 \\
    		\midrule
                w/o ID loss & 6.598 & 0.929 \\
                w. ID loss $\times$ 2.0 & 7.010 & 0.951 \\
    		\bottomrule[1.0pt]
    	\end{tabular}
    \end{minipage} &
    \begin{minipage}[t]{.48\linewidth}
            \centering
            \scriptsize
            \captionof{table}{\textbf{Ablation study on reference design in appearance generation model.} See Sec.~\ref{ablation_study} for details.}
            \label{tab:VAE_ablation}
            \resizebox{\columnwidth}{!}{
            \begin{tabular}{l|cc}
    		\toprule[1.0pt]
    		Method & PSNR$\uparrow$ & FID$\downarrow$ \\
                \midrule
                Ours (masked multi-ref.) & 34.153 & 4.224 \\
    		\midrule
                w. full single-ref. & 31.750 & 6.62 \\
    		  w. masked single-ref. & 32.759 & 6.453 \\
    		\bottomrule[1.0pt]
    	\end{tabular}
            }
        \end{minipage}
        \vspace{0mm}
        \end{tabular}
\end{table}

We conduct thorough ablation studies on the technical designs of our motion generation model and appearance generation model. 

\subsubsection{Condition Types in Motion Generation Model.}

Given that the model cannot successfully predict the entire facial landmarks conditioned solely on the upper face landmarks and speech input, we provide the model with the identity embedding of a reference frame for identity guidance. To demonstrate the validity of our reference information conditioning, we conduct a series of experiments on different condition types within our diffusion model. Specifically, we further explore three distinct conditioning strategies, including (1) concatenating the key landmark with input; (2) adding the key landmark feature to the input; and (3) integrating the key landmark via cross-attention. 

In Tab.~\ref{tab:diff_ablation}, our conditioning design outperforms others on both lip-sync quality and identity (ID) similarity. With other conditioning strategies, the model struggles to excel in both lip-sync quality and ID similarity, owing to the lack of constraint for identity learning. In contrast, our approach directly provides the model with identity features via our identity extractor. 

\subsubsection{Identity Loss in Motion Generation Model.}
To evaluate the effectiveness of our identity (ID) loss design, we further perform experiments using two different loss configurations: (1) excluding ID loss, and (2) incorporating ID loss with a weight of 2.0. As shown in Tab.~\ref{tab:loss_ablation}, incorporating the ID loss significantly enhances the ID preservation. However, compared to MyTalk which uses the ID loss weight of $1.0$, the higher ID loss weight compromises the model performance in lip sync.

\subsubsection{Reference Design in Appearance Generation Model.} 
We conducted experiments using three different reference configurations within our appearance generation model in Table~\ref{tab:VAE_ablation}. The first configuration uses a single image as the reference (full single-ref.). The second configuration employs a single image with non-lip region masked out (masked single-ref.). Finally, our MyTalk model implements a multi-reference setup that involves three masked reference images. The results in Table~\ref{tab:VAE_ablation} indicate that masking operation helps in image quality and our multi-reference strategy further enhances the appearance generation.

\subsection{Controllable Generation}
\begin{figure}[t]
  \centering
  \includegraphics[height=4.7cm]{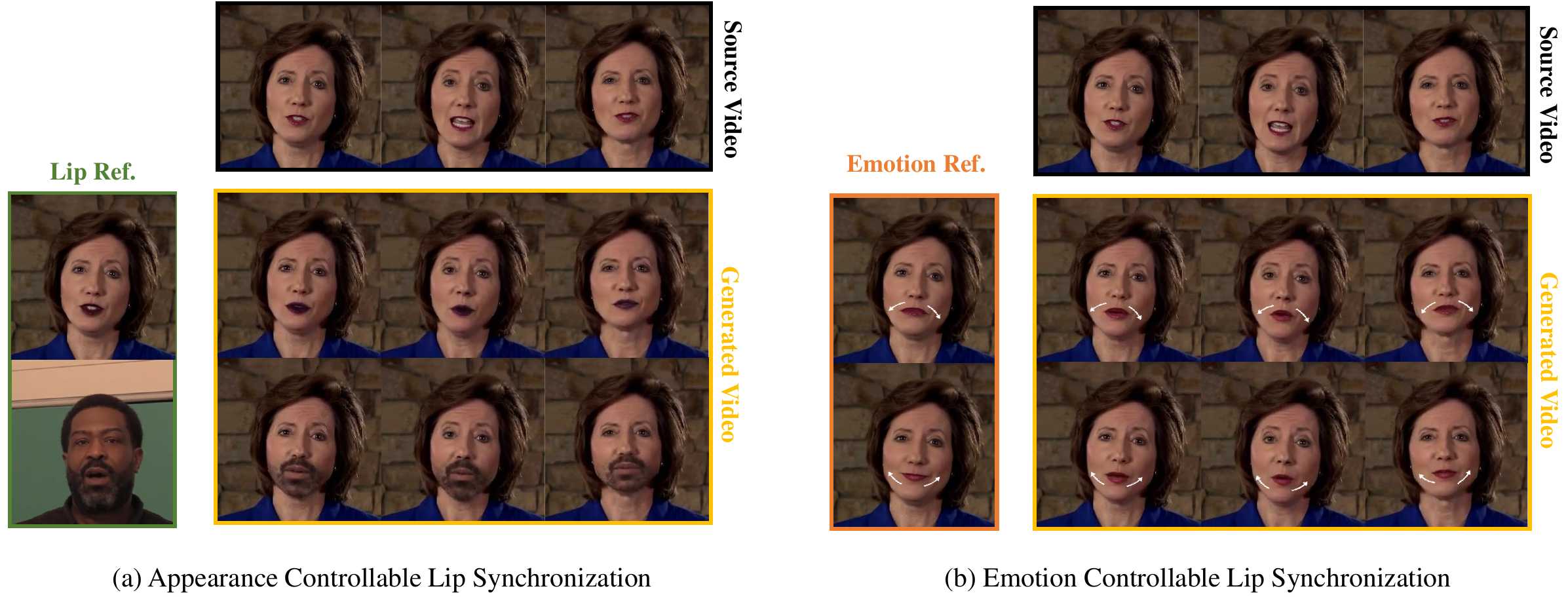}
  \caption{Examples of controllable generation. Benefiting from our disentanglement and identity preservation designs, MyTalk shows novel properties in controlling appearance and emotion while editing the talking video.}
  \label{fig3:appearance_edit}
\end{figure}

Benefiting from our disentangled modeling, MyTalk can accurately integrate the appearance and motion conditions into generated videos as separate factors, which enables us to perform intriguing manipulations during the generation process. As shown in Fig.~\ref{fig3:appearance_edit} (a), we can edit the lip region appearance by providing the model with a variety of reference images. 
In addition, we can emotionally control the talking video by using the emotion reference to guide both motion and appearance generation, see Fig.~\ref{fig3:appearance_edit} (b) for visualization.

\section{Conclusion}
We propose to disentangle the motion and appearance in talking video lip sync and further divide the task into two sub-problems: speech-driven lip motion generation and visual appearance synthesis. We introduce MyTalk, a two-stage model, to separately generate the motion and appearance. Specifically, we utilize a speech-driven diffusion model to generate the motion, and propose an identity loss to preserve the identity. In the second stage, we introduce a motion-conditioned appearance generation model, which separately encodes the lip, non-lip appearance and motion, and then integrates them with a learned fusion module. Extensive experiments demonstrate the superiority of our method in visual detail preservation and lip-sync quality. Benefiting from the scaled training and disentangled modeling, MyTalk generalizes well to out-of-domain characters and shows great controllability to lip appearance and talking emotions. See the Appendix for more discussion on limitations and future work.

\bibliographystyle{splncs04}
\bibliography{main}

\clearpage

\appendix
\section*{Appendix}
We organize our supplementary material as follows:
\begin{itemize}
    \item Sec.~\ref{sec1} gives qualitative comparisons with the baseline models mentioned in the main text.
    \item Sec.~\ref{sec2} provides some additional details on the workflow and architecture of the framework, its configuration, and dataset collection. 
    \item Sec.~\ref{sec3} demonstrates the limitations and future work of our proposed MyTalk.
\end{itemize}

\section{Qualitative Comparison}
\label{sec1}
\begin{figure}[!htbp]
  \centering
  \includegraphics[height=8.5cm]{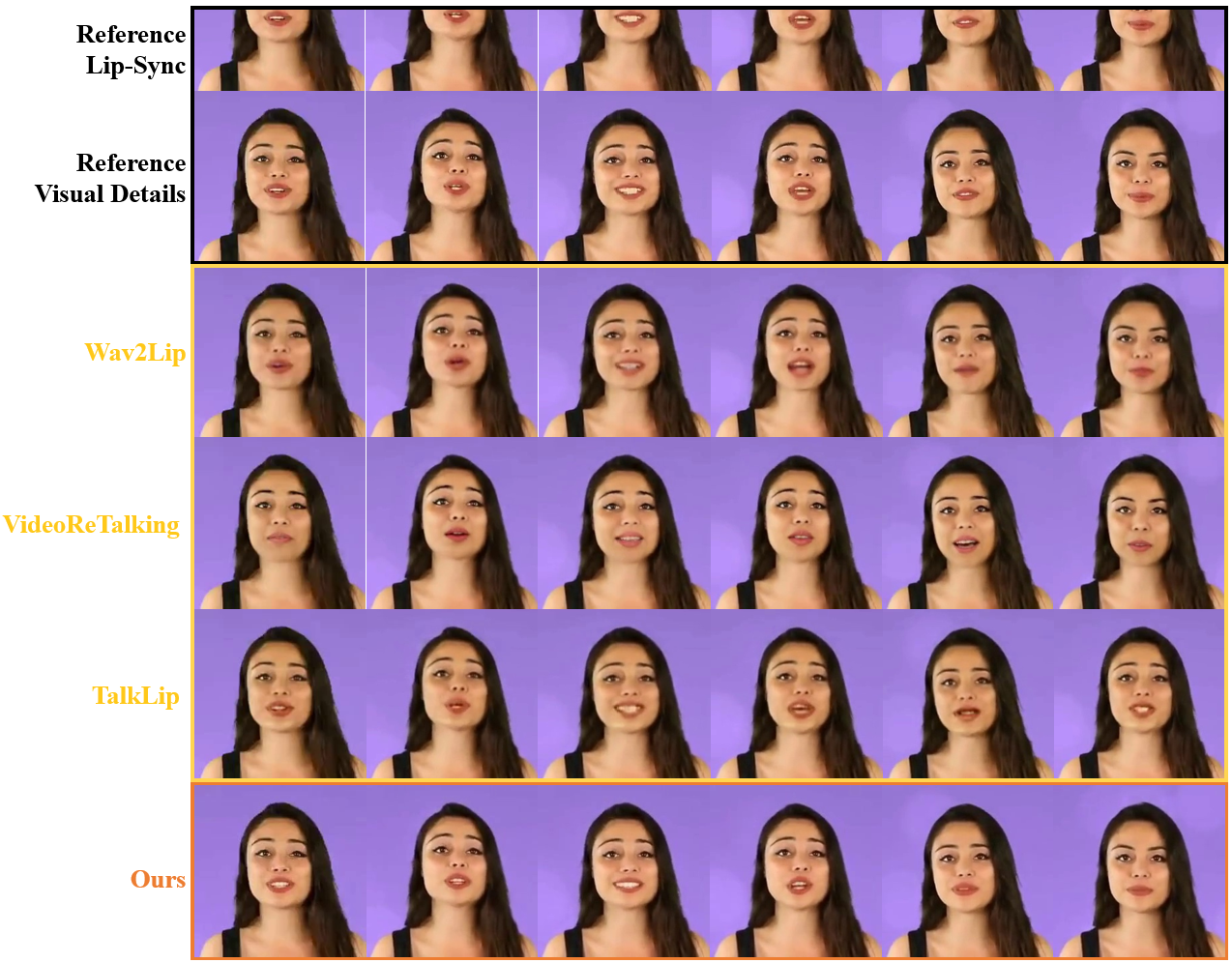}
  \caption{Qualitative comparison on paired video-speech data.}
  \vspace{-10pt}
  \label{fig3:paired}
\end{figure}

\begin{figure}[!htbp]
  \centering
  \includegraphics[height=8.5cm]{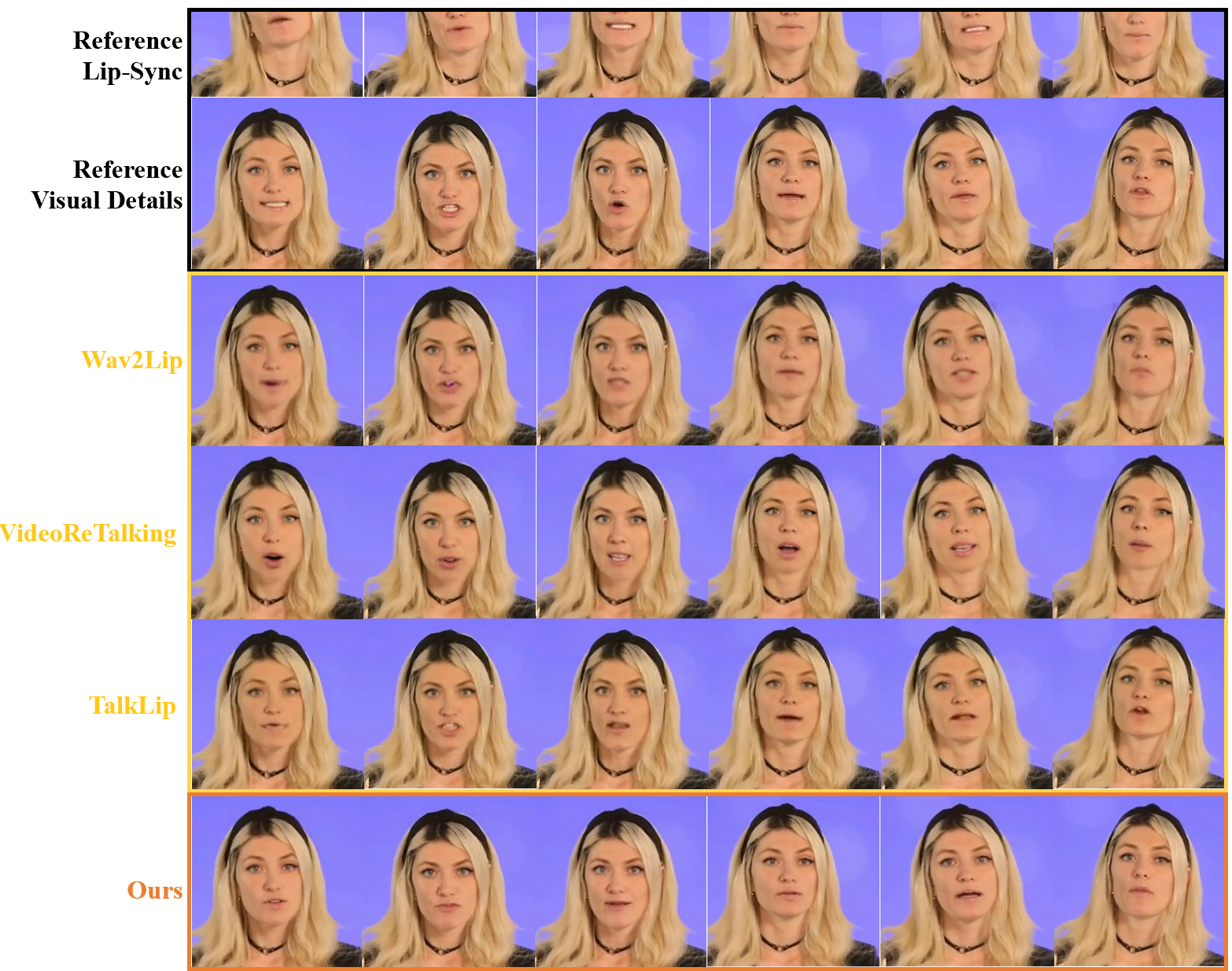}
  \vspace{-3pt}
  \caption{Qualitative comparison on unpaired video-speech data, wherein the video and speech are sampled from the same identity and different segments.}
  \vspace{-2pt}
  \label{fig4:unpaired}
\end{figure}

\begin{figure}[!htbp]
  \centering
  \includegraphics[height=8.5cm]{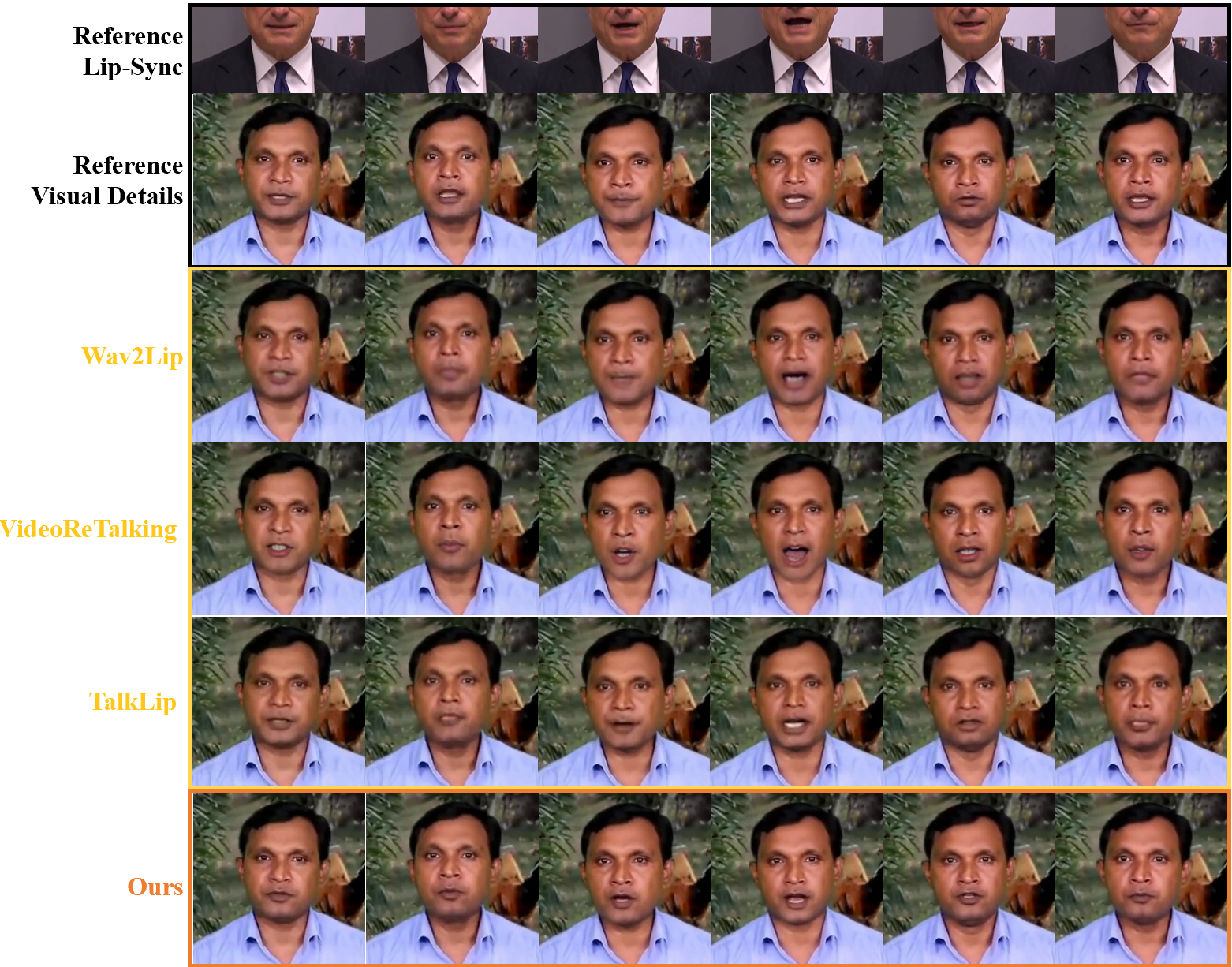}
  \vspace{-3pt}
  \caption{Qualitative comparison on unpaired video-speech data, wherein the video and speech are sampled from different identities and segments.}
  \label{fig5:unpaired}
\end{figure}

We provide additional visualizations on qualitative comparisons to supplement the main paper. Fig.~\ref{fig3:paired}, Fig.~\ref{fig4:unpaired} and Fig.~\ref{fig5:unpaired} show the generation results on paired video-speech data, unpaired data from the same identity and unpaired data from different identities, respectively. The visualization comparisons all demonstrate our superiority in visual detail preservation and lip-sync quality.



\clearpage

\section{Additional Dataset and Framework Details}
\label{sec2}
\subsection{Framework Details}
\subsubsection{Speech-driven Motion Generation Model}
As demonstrated in Sec.~3.3 of the main text, we adopt a diffusion model to generate the landmark motion sequences. It takes the conditions $C=\{t, z^{s}, z^{id}\}$, where $t$ is the diffusion time step, $z^{s}$ is the speech feature, and $z^{id}$ is the identity feature gotten by our identity extractor. The model design is shown in Fig.~3(a) of the main text, and the expanded details and the inference process are presented as follows. 

\begin{figure}[hb]
  \centering
  \includegraphics[height=8.8cm]{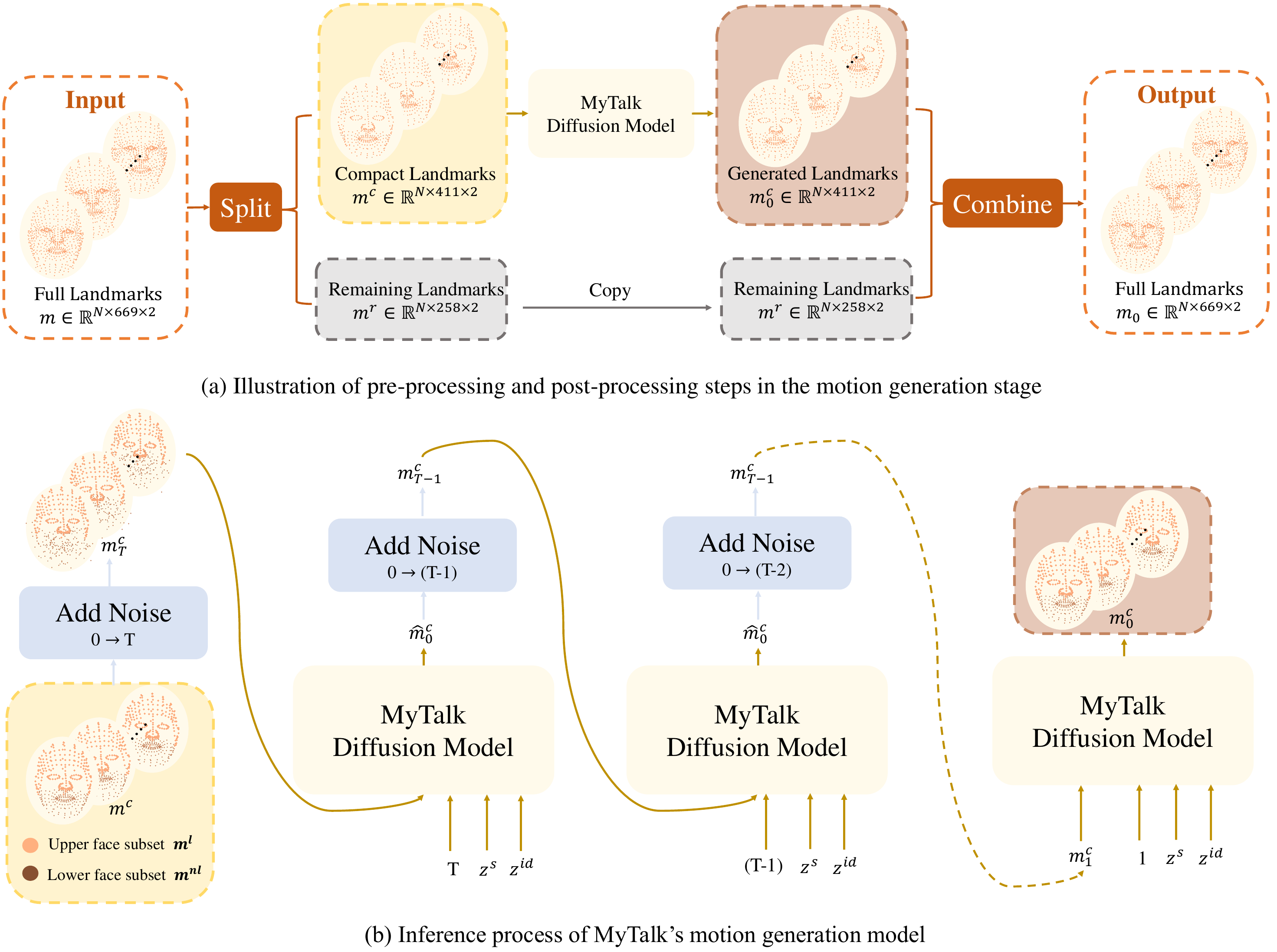}
  \caption{Framework workflow of MyTalk's motion generation model.}
  \label{fig1:diff_inference}
\end{figure}

As illustrated in Fig.~\ref{fig1:diff_inference} (a), we introduce some processes beyond the diffusion. Specifically, we divide the full landmarks $m\in\mathbb{R}^{N\times 669\times 2}$ into two parts, the compact landmarks $m^c\in\mathbb{R}^{N\times 411\times 2}$ and the remains $m^c\in\mathbb{R}^{N\times 258\times 2}$. Here, the numbers 669, 411 and 258 represent the number of key points within each set of landmarks, with each point described by two-dimensional coordinates for its horizontal and vertical positions.
This division is based on our empirical observations, which reveal that the remaining landmarks $m^r$ contain unnecessary information for speech-related motion generation.    
Subsequently, as shown in Fig.~\ref{fig1:diff_inference} (b), the diffusion process is conducted on the compact landmark $m^c$. The generated landmarks $m^c_0$ are then combined with the remaining landmarks $m^r$ to form the final output $m_0$. 

\subsubsection{Motion-conditioned Appearance Generation Model}
\begin{figure}[ht]
  \centering
  \includegraphics[height=5.5cm]{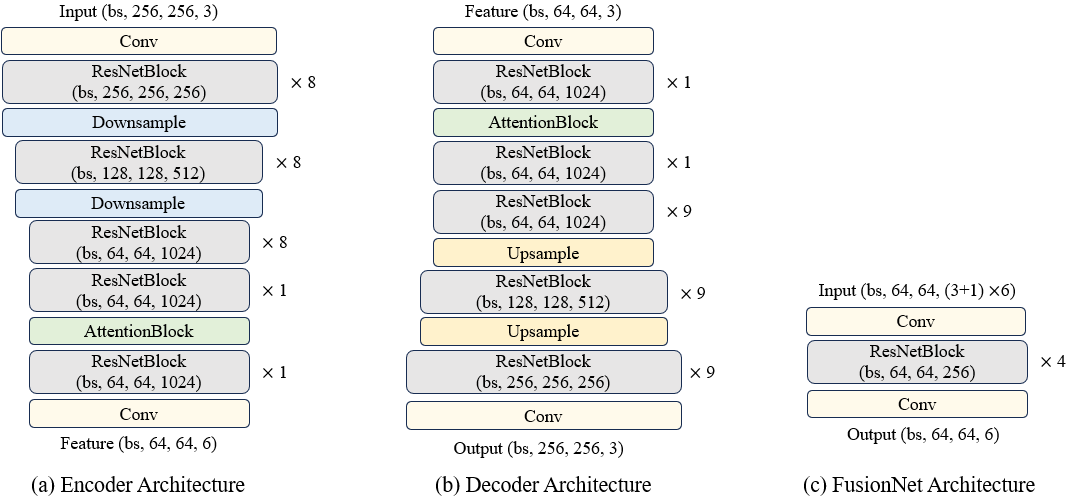}
  \caption{Framework architecture of MyTalk's appearance generation model.}
  \vspace{-5mm}
  \label{fig2:VAE}
\end{figure}

Our appearance generation model consists of three encoders, a FusionNet and one decoder, which is briefly illustrated in Fig.~3 (c). These encoders adopt the same architectures and the detailed architecture design is depicted in Fig.~\ref{fig2:VAE} (a). The symmetrical architecture of the decoder is shown in Fig.~\ref{fig2:VAE} (b). As illustrated in in Fig.~\ref{fig2:VAE} (c), we employ a FusionNet to fuse the lip and non-lip appearance features. This network takes the concatenated lip appearance feature and three non-lip appearance features as input and outputs the blended appearance feature. 

\subsubsection{Discussion on the Use of Landmark}
Like previous research~\cite{zhang2023sadtalker},
an alternative approach to motion generation is predicting the coefficients of a 3D Morphable Model (3DMM)~\cite{blanz2023morphable}. However, we have observed in practice that it is challenging to obtain these coefficients for out-of-domain characters, especially when dealing with cartoon-based ones. To realize more generalized lip synchronization, we employ more effective and robust landmarks to represent the motion.

\subsection{Dataset Details}
\label{dataset_detail}
As demonstrated in Sec. 4.1 of the main text, we assemble a large dataset comprising data from both public sources and internally collected datasets. Our appearance generation model is trained on the entire dataset, whereas our motion generation model adopts a subset of it. The high-quality subset excludes the parts with no speech, unaligned speech, and poor lip-syncing. The details are illustrated in Tab.~\ref{tab:dataset}.

\begin{table}[th]
  \caption{Statistics of our dataset.}
  \label{tab:dataset}
  \centering
  \begin{tabular}{c@{\hspace{5mm}}c@{\hspace{5mm}}c}
    \toprule
    Tasks & Time Duration & Number of Identity \\
    \hline
    Appear. Generation & 1,169 hours & 15,969 \\
    Motion Generation & 148 hours & 1,027 \\
    \bottomrule
  \end{tabular}
\end{table}

\subsection{Configuration}
The model size and hyper-parameters of each component are detailed in Tab.~\ref{tab:configuration}.

\vspace{-4mm}
\setlength{\extrarowheight}{1pt}
\begin{table}[!htbp]
  \caption{Configuration of MyTalk.}
  \vspace{-2mm}
  \label{tab:configuration}
  \centering
  \begin{tabular}{@{\hspace{2mm}}c@{\hspace{1mm}}|@{\hspace{1mm}}c@{\hspace{1mm}}|c|@{\hspace{1mm}}c}
    \toprule
    Model & Component & Model Size & Hyper-parameters \\
    \hline
    \multirow{12}{*}{\makecell[c]{Motion\\ Generation \\ Model}} & Audio Encoder & 24.6M & \makecell[l]{Conformer Layers 4 \\ Hidden Dim. 768 \\ FFN Dim 2048 \\ Attention Heads 8 \\ Kernel Size 13 \\ Dropout 0.1} \\
    \cline{2-4}
    & Audio Downsample & 786K & \makecell[l]{Kernel Size 3 \\ Stride 2 \\ Padding 1} \\
    \cline{2-4}
    & Identity Extractor & 4.3M &  \makecell[l]{MLP Layers 4 \\ Hidden Dim 768} \\
    \cline{2-4}
    & Conformer Backbone & 188M &  \makecell[l]{Conformer Layers 12 \\ Hidden Dim. 768 \\ FFN Dim 2048 \\ Attention Heads 12 \\ Kernel Size 5 \\ Dropout 0.1 } \\
    \hline
    \multirow{15}{*}{\makecell[c]{Appearance\\ Generation \\ Model}} & Lip Appear. Encoder & \multirow{3}{*}{237M} & \multirow{3}{*}{\makecell[l]{Hidden Dim. 256 \\ ResnetBlock Layers 8 \\ Kernel Size 3}} \\
    & Non-lip Appear. Encoder &  & \\ 
    & Landmark Encoder &  &  \\ 
    \cline{2-4}
    & FusionNet & 4.8M & \makecell[l]{Hidden Dim. 256\\ ResnetBlock Layers 4 \\ Kernel Size 3} \\ 
    \cline{2-4}
    & Appear. Quant. Conv. & 42 & \multirow{4}{*}{\makebox[15pt][r]{Kernel Size 1}} \\
    & Landmark Quant. Conv. & 42 & \\
    & Appear.-Landmark Quant. Conv. & 21 & \\
    & Post-quant. Conv. & 12 & \\
    \cline{2-4}
    & Decoder & 280M & \makecell[l]{Hidden Dim. 256 \\ ResnetBlock Layers 8 \\ Kernel Size 3} \\ 
    \cline{2-4}
    & Discriminator & 17.5M & \makecell[l]{KL Loss Weight 1.e$^{-6}$ \\ Dis. Loss Weight 0.5}\\
    \bottomrule
  \end{tabular}
\end{table}
\clearpage

\section{Limitations and Future Work}
\label{sec3}
Firstly, as stated in Sec.~\ref{dataset_detail}, the speech-to-motion generation training dataset is currently limited in size. We plan to expand this dataset to enhance the lip-sync quality and explore the scalability of our method.
Secondly, as detailed in Table~\ref{tab:method_compare}, our method requires more inference time due to the two-stage modeling. However, MyTalk outperformes other methods in terms of visual quality (PSNR, FID) and ID similarity (ID$_{sim}$) with a noticeable performance gap, which justifies extra time investment required by MyTalk.
Finally, the current framework leverages a specific pre-trained landmark extractor~\cite{wood2021fake}, which may hinder the end-to-end learning of our model. We leave the effectiveness evaluation of our method on alternative types of landmarks (e.g., sparse landmark detected by~\cite{kazemi2014one}) and no landmarks (e.g., disentangle the motion and appearance without the help of landmarks) for future work. 

\begin{table}[h]
  \caption{Comparisons with one-stage models.}
  \vspace{-2mm}
  \label{tab:method_compare}
  \centering
  \begin{tabular}{l|ccc|cc|cc}
    \toprule
    \multirow{2}{*}{Method} & \multicolumn{5}{c|}{Metric Evaluation} & \multicolumn{2}{c}{Model Efficiency} \\
    \cline{2-8}
    & PSNR$\uparrow$ & FID$\downarrow$ & ID$_{sim}\uparrow$ & Sync$_{dist}\downarrow$ & Sync$_{conf}\uparrow$ & \#param.(B) & Time(s) \\
    \hline
    Wav2Lip~\cite{prajwal2020lip} & 30.967 & 7.703  & 0.921 & \textbf{6.249} & \textbf{7.829} & 0.14 & \textbf{32.392} \\
    VideoReTalking~\cite{cheng2022videoretalking} & 29.628 & 9.628 & 0.895 & 6.978 & 6.986 & 0.69 & 897.671 \\
    TalkLip~\cite{wang2023seeing} & \underline{31.140} & 7.401 & 0.926 & 8.107 & 5.300 & 0.53 & 67.756\\
    \hline
    MyTalk(ours) & \textbf{33.547} & \textbf{5.142} & \textbf{0.947} & \underline{6.544} & \underline{7.461} & 1.23 & 631.497 \\
    \bottomrule
  \end{tabular}
\end{table}

\end{document}